# Not All Unlabeled Data Are Equal: Unlabeled Data Selection for Semi-Supervised QSAR Modeling with Large Chemical Repositories

Oliver Majer[1,2], Sašo Džeroski[1] [0000−0003−2363−712X], and Jurica Levatić[1] [0000−0002−4025−051X]

[1] Jožef Stefan Institute, Jamova cesta 39, Ljubljana, Slovenia
{jurica.levatic,saso.dzeroski}@ijs.si
[2] University of Ljubljana, Faculty of Computer and Information Science, Večna pot 113, Ljubljana, Slovenia
olivermajer7@gmail.com

**Abstract.** Semi-supervised learning (SSL) is particularly attractive in domains where labels are expensive to obtain, while unlabeled examples are readily available. Quantitative Structure–Activity Relationship (QSAR) modelling, which predicts molecular biological activity from chemical structure, is one such domain. Obtaining activity measurements requires costly wet-lab experiments, resulting in small labeled datasets, while chemical repositories such as ChEMBL contain millions of unlabeled molecules. Despite this apparent suitability, the potential of large chemical databases to improve QSAR models via SSL remains largely unexplored. Furthermore, it is unclear how unlabeled data should be selected in realistic SSL settings with genuine unlabeled data.

We conduct a systematic study of unlabeled-data selection using 10 QSAR datasets extracted from ChEMBL and a pool of more than 400,000 unlabeled molecules. To this end, we evaluate three established SSL algorithms (self-training, co-training, and label propagation) and formulate four unlabeled-data selection strategies based on random, nearest-neighbour, furthest-neighbour, and mixed sampling.

Our results show that unlabeled data from large chemical repositories can improve QSAR models, with SSL outperforming the supervised baseline on all 10 datasets. Moreover, the most reliable improvements are generally achieved by selecting a relatively small number of chemically distant molecules. This suggests that the composition of the unlabeled set may be more important than its size, challenging the common premise that SSL necessitates an abundance of unlabeled data.

## 1 Introduction

Semi-supervised learning (SSL) aims to improve predictive performance by leveraging both labeled and unlabeled data during model training [3]. Unlike supervised learning, which relies exclusively on labeled examples, SSL exploits abundant unlabeled data to learn more accurate models. SSL has been successfully applied across many domains, including computer vision [4], natural language processing [19], and bioinformatics [13], particularly in settings where obtaining labels is expensive or time-consuming. Common SSL approaches include self-training, co-training, consistency regularization, and graph-based methods [5].

Quantitative Structure–Activity Relationship (QSAR) modeling represents an attractive application domain for SSL. QSAR modeling aims to establish relationships between the structural and physicochemical properties of molecules and their biological activities using computational models. By predicting molecular properties, QSAR models can accelerate drug discovery and substantially reduce development costs [12]. However, the predictive performance of QSAR models is often limited by the availability of labeled data [8].

In QSAR studies, labeled datasets are typically limited in size because biological activity measurements require costly and time-consuming wet-lab experiments. As a result, many publicly available QSAR datasets contain only tens to hundreds of molecules (Figure 1). On the other hand, large public chemical repositories such as ChEMBL contain millions of compounds, the vast majority of which lack labels for a specific prediction task [7]. This combination of scarce labeled data and abundant unlabeled molecules closely matches the premise underlying SSL and offers a potential for exploiting large chemical databases to improve QSAR models. However, despite the apparent suitability of QSAR modelling for semi-supervised learning, the potential of large chemical repositories to improve QSAR models within the SSL setting remains largely unexplored.

Previous studies leveraged large chemical repositories to learn molecular representations and enhance QSAR models through self-supervised pretraining followed by task-specific fine-tuning [9, 11]. Unlike self-supervised learning, which relies on computationally expensive pretraining to learn general-purpose molecular representations, classical SSL incorporates unlabeled molecules directly into the training of the target QSAR model. As a result, SSL can be applied in a task-specific manner without requiring large-scale pretraining and can be readily deployed whenever a small labeled dataset and a large pool of unlabeled molecules are available. Furthermore, because unlabeled data are exploited in a task-specific manner, classical SSL can capture information relevant to the target prediction task that may not be fully reflected in general-purpose molecular representations.

A closely related issue within the SSL field concerns the selection of unlabeled data. Namely, the vast majority of studies evaluate SSL methods using simulated unlabeled data generated by removing labels from a portion of the existing labeled dataset. This setup implicitly assumes that the unlabeled and labeled data come from the same distribution and typically deals with a relatively small pool of unlabeled data (i.e., a part of labeled dataset), which is unselectively provided to SSL algorithms. This setting, however, does not correspond to the real-world scenario which inspired the development of SSL in the first place: scarce labeled examples accompanied by a vast pool of unlabeled data (with a possible distribution mismatch). It is therefore currently unclear how unlabeled data should be selected in real-world SSL applications with access to large pools of genuinely unlabeled data, and how different selection strategies affect the predictive performance of SSL algorithms.

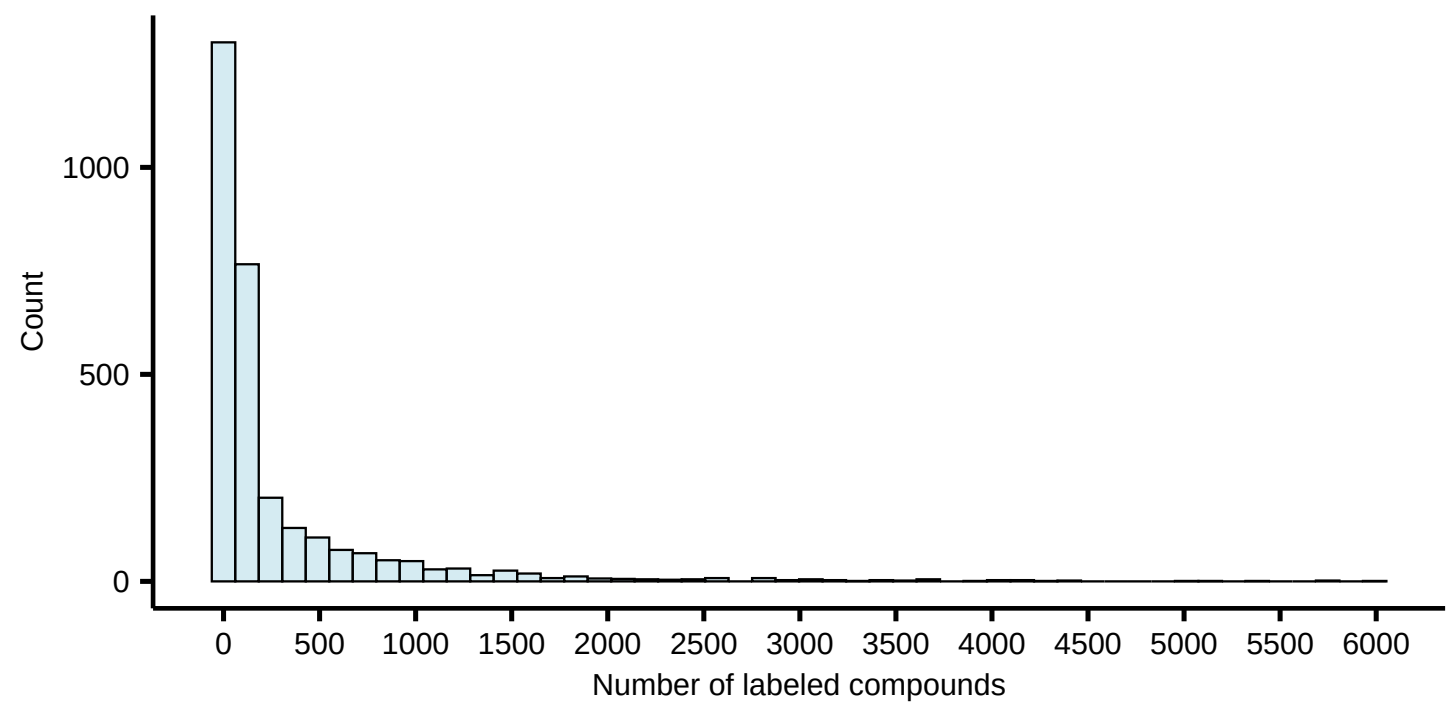

Fig. 1: Histogram of dataset sizes (in terms of number of labelled compounds) for 2970 biological targets extracted from the ChEMBL database.

To this end, we conduct a systematic study of unlabeled data selection in QSAR modeling using 10 datasets extracted from the ChEMBL database and a pool of more than 400,000 unlabeled molecules. With this, we simultaneously address the two aforementioned open questions: (i) whether large chemical databases can be exploited to enhance QSAR models via SSL, and (ii) how unlabeled data selection affects the performance of SSL models in a real-world setting. We consider three well-known SSL methods (*self-training* [19], *co-training regression* [1], and *label propagation* [22]) and propose four different unlabeled data selection strategies: random sampling, nearest-neighbour selection, furthest-neighbour selection, and mixed strategy. We evaluate these strategies across varying unlabeled set sizes.

Our experiments reveal a somewhat counter-intuitive result: selecting a relatively small number of chemically distant molecules yields more reliable improvements in SSL predictive performance. This finding challenges the typical SSL premise of exploiting large amounts of unlabeled data and suggests that the composition of the unlabeled set may be more important than its size. From a practical perspective, it further indicates that better SSL performance may be achievable using only a small amount of unlabeled data, potentially reducing the computational cost of model training.

## 2 Methods

This section presents the SSL methods used in our experiments and the unlabeled-data selection strategies considered in this study. The associated Python implementation is available at: https://github.com/geekeemoli/Semi-supervised-learning-unlabeled-data-selection.

### 2.1 Semi-supervised algorithms

We consider three SSL methods to predict the bioactivity of molecules: *self-training* [19], *co-training regression* [1], and *label propagation* [22]. All three are *wrapper*-style methods that augment a small labeled set $\mathcal{L} = \{(\mathbf{x}_i, y_i)\}_{i=1}^{l}$ with pseudo-labels inferred for an unlabeled pool $\mathcal{U} = \{\mathbf{x}_j\}_{j=l+1}^{l+u}$, where $y_i \in \mathbb{R}$ denotes the continuous target. As the base learner we use a random forest (RF) regressor [2], which provides both a prediction and an estimate of prediction uncertainty. Below, we describe each method in brief, while full details can be found in the original publications.

**Self-training** A model is first trained on the labeled data and then used to predict the unlabeled examples; the most confident predictions are added to the training set as pseudo-labels, and the process is repeated. Since scikit-learn's `SelfTrainingClassifier` supports only classification, we implemented a regression variant of self-training using scikit-learn's random forest [15], in which confidence is derived from the variance among the predictions of the individual trees in the forest.

Concretely, for an unlabeled example $\mathbf{x}$, the forest of $T$ trees $\{h_t\}_{t=1}^{T}$ yields as a prediction the mean of the predictions of the $T$ trees $\bar{y}(\mathbf{x})$ and standard deviation $s(\mathbf{x})$:

$$\bar{y}(\mathbf{x}) = \frac{1}{T}\sum_{t=1}^{T} h_t(\mathbf{x}), \qquad s(\mathbf{x}) = \sqrt{\frac{1}{T}\sum_{t=1}^{T} \big(h_t(\mathbf{x}) - \bar{y}(\mathbf{x})\big)^2}, \tag{1}$$

where a low $s(\mathbf{x})$ indicates that the trees agree and the prediction is reliable. At each iteration, every remaining unlabeled example satisfying $s(\mathbf{x}) \leq \theta$ is pseudo-labeled with $\bar{y}(\mathbf{x})$, added to the labeled set, and removed from the unlabeled pool; the forest is then retrained. The confidence threshold $\theta$ controls how conservative the pseudo-labeling is (smaller $\theta$ admits fewer, more certain examples). The procedure runs for a fixed number of iterations or until no example is added.

**Co-training** Co-training [1] trains two models on distinct *views* of the data and lets each model teach the other: examples that one model confidently labels are added to the training set of its peer. Originally proposed for classification with two sufficient and redundant feature views, co-training was later extended to regression by Zhou and Li [21] in their CoReg algorithm. Notably, CoReg dispenses the multi-view assumption (which rarely holds for real regression data) and instead diversifies its two learners by using $k$-nearest-neighbor regressors with different distance orders on the *full* feature set.

We adopt the co-training *scheme* for regression but, following the original multi-view formulation [1], derive the two learners from a feature split rather than from different distance metrics: the 1024-bit fingerprint is partitioned into two disjoint halves, $\mathbf{x} = (\mathbf{x}^{(1)}, \mathbf{x}^{(2)})$, and two random forest regressors $h_1$ and $h_2$ are trained, one on each view. At every iteration, both regressors predict the remaining unlabeled examples, and the examples on which the two views *agree most* are selected, i.e., those that minimize the squared disagreement

$$\Delta(\mathbf{x}) = \big(h_1(\mathbf{x}^{(1)}) - h_2(\mathbf{x}^{(2)})\big)^2. \tag{2}$$

The $k$ examples with the smallest $\Delta(\mathbf{x})$ are pseudo-labeled and exchanged across views: each example is added to the training set of $h_1$ with the label predicted by $h_2$, and to the training set of $h_2$ with the label predicted by $h_1$. Both regressors are then retrained on their enlarged labeled sets. After a fixed number of iterations, predictions are obtained from the regressor trained on the first view.

**Graph propagation** Label propagation is a graph-based, transductive SSL method that spreads label information from labeled to unlabeled examples through a similarity graph [22, 20]. Each example is a node, and edges encode pairwise similarity; the method rests on the smoothness/manifold assumption that strongly connected nodes should receive similar labels. Because descriptive attributes are binary molecular fingerprints (Section 3.1), we build the graph from the *Tanimoto* (Jaccard) similarity [16],

$$T(\mathbf{x}_a, \mathbf{x}_b) = \frac{\mathbf{x}_a \cdot \mathbf{x}_b}{\|\mathbf{x}_a\|^2 + \|\mathbf{x}_b\|^2 - \mathbf{x}_a \cdot \mathbf{x}_b}, \tag{3}$$

which is the standard measure of chemical similarity for fingerprints. We retain the $k$ nearest neighbors of every node and row-normalize the resulting affinity matrix to obtain a stochastic transition matrix $P$, partitioned into labeled ($l$) and unlabeled ($u$) blocks.

Following the harmonic-function formulation of Zhu and Ghahramani [22], the unlabeled scores $\mathbf{f}_u \in \mathbb{R}^u$ are obtained by iterating the propagation update

$$\mathbf{f}_u^{(t+1)} = P_{uu}\,\mathbf{f}_u^{(t)} + P_{ul}\,\mathbf{y}_l, \tag{4}$$

clamping the labeled values $\mathbf{y}_l$, until convergence (or a maximum number of iterations). The unlabeled scores are initialized to the mean of the labeled targets. To avoid injecting uninformative pseudo-labels, only those unlabeled examples whose propagated value deviates from this prior by more than a fixed margin are retained as pseudo-labels. The labeled set augmented with these pseudo-labels is then used to train a final random forest regressor that produces the predictions on the test set, yielding an inductive model rather than a transductive one.

## 2.2 Unlabeled data selection strategies

A central question of this study is how the composition and size of the unlabeled set affect SSL performance. To this end, we formulate four unlabeled data selection strategies: *Random*, *Nearest*, *Furthest*, and *Mixed*. Before describing these strategies, we introduce the common setting in which they operate.

Given a labeled training set $\mathcal{L}$ of size $n$ and a large unlabeled pool $\mathcal{U}$, each strategy selects a budget of $m$ molecules from $\mathcal{U}$ to serve as the unlabeled data for the SSL algorithms. To study how the amount of unlabeled data affects the SSL performance, we consider unlabeled-set sizes $m \in \{100, 250, 500, 1000, 2500, 5000\}$. Distances between molecules are measured with the *Tanimoto* (Jaccard) distance on their 1024-bit binary molecular fingerprint representation (Section 3.1), $d(\mathbf{x}_a, \mathbf{x}_b) = 1 - T(\mathbf{x}_a, \mathbf{x}_b)$, where $T$ is the Tanimoto similarity of Eq. (3).

**Random** As a baseline, $m$ molecules are drawn at random from $\mathcal{U}$, without repetition and independently of the labeled set. This strategy ignores the structure of the chemical space and isolates the effect of simply increasing the quantity of unlabeled data, against which the targeted strategies below are compared.

**Nearest** This strategy populates the unlabeled set with molecules that are chemically *similar* to the labeled examples. To spread the budget across the whole labeled space rather than concentrating it around a few molecules, we use a *quota* scheme: each of the $n$ labeled molecules is assigned a quota of

$$k = \left\lceil \frac{m}{n} \right\rceil \tag{5}$$

unlabeled neighbors. The labeled molecules are processed in a random (seed-fixed) order, and each one claims its $k$ nearest unlabeled neighbors that have not yet been claimed by another labeled molecule; if its closest neighbors are already taken, it falls back to the next-nearest available ones. Randomizing the order prevents the first labeled molecules from monopolizing the most attractive neighbors and keeps the selection reproducible across folds.

**Furthest** This strategy is the complement of the previous one: using the same quota scheme of Eq. (5) and random processing order, each labeled molecule claims its $k$ *furthest* unclaimed unlabeled neighbors instead of its nearest ones. The selected molecules therefore lie in regions of the chemical space that are dissimilar to the labeled examples, probing whether unlabeled data that extends, rather than reinforces, the coverage of the labeled set is beneficial or detrimental to the SSL models.

**Mixed** The mixed strategy combines exploitation and exploration by splitting each labeled molecule's quota evenly between its nearest and furthest neighbors. Of the $k$ slots in Eq. (5), $\lceil k/2 \rceil$ are filled with the nearest unclaimed neighbors and $\lfloor k/2 \rfloor$ with the furthest unclaimed neighbors (an odd remainder is assigned to the nearest half). This yields an unlabeled set that simultaneously reinforces the immediate neighborhood of the labeled molecules and samples the more distant parts of the chemical space.

Table 1: Characteristics of the 10 QSAR datasets extracted from the ChEMBL database. All datasets use 1024-bit FCFP4 fingerprints as descriptors. ChEMBL target names correspond to the human protein target.

| ChEMBL ID | Target name | #Molecules | #Features |
|---|---|---|---|
| CHEMBL1907606 | Mitogen-activated protein kinase ERK1/ERK2 | 30 | 1024 |
| CHEMBL4010 | Capsid scaffolding protein | 33 | 1024 |
| CHEMBL4835 | L-lactate dehydrogenase A chain | 35 | 1024 |
| CHEMBL4208 | Proteasome component C5 | 37 | 1024 |
| CHEMBL2457 | Lipoxygenase ALOX15B | 70 | 1024 |
| CHEMBL5281 | Cytochrome P450 2E1 | 75 | 1024 |
| CHEMBL5685 | Solute carrier family 22 member 1 | 75 | 1024 |
| CHEMBL3535 | cGMP-specific phosphodiesterase 9A | 112 | 1024 |
| CHEMBL4698 | Malonyl-CoA decarboxylase, mitochondrial | 264 | 1024 |
| CHEMBL2434 | C-X-C chemokine receptor type 2 | 525 | 1024 |

# 3 Experimental design

## 3.1 Data

We selected the datasets from the large-scale QSAR collection that Olier et al. [14] compiled from the ChEMBL database [6], as distributed through the OpenML platform [18]. The version we use comprises 2,970 single-target QSAR datasets[1]. Each molecule is represented by a 1024-bit FCFP4 fingerprint [14] encoding the presence and absence of molecular substructures, and is associated with a curated numeric bioactivity value (pXC50) against a specific protein target.[2]

From this collection we selected 10 datasets for our experiments. The selection was random, but constrained so as to span a broad range of labeled-set sizes: the chosen datasets range from only 30 labeled molecules (`CHEMBL1907606`) to 525 (`CHEMBL2434`), and include small, intermediate, and comparatively large QSAR tasks. This diversity lets us assess how the semi-supervised methods and the unlabeled-data selection strategies behave as the amount of labeled data varies. The characteristics of the 10 datasets are summarized in Table 1.

The pool of unlabeled data is formed from the union of all molecules across the full collection of 2,970 QSAR datasets. After removing duplicate compounds shared by multiple targets, this yields a pool of 415,654 unique molecules, each represented by the same 1024-bit FCFP4 fingerprint. For each labeled dataset, this pool (excluding the molecules labeled for the target under consideration) acts as the external source of unlabeled examples from which the selection strategies of Section 2.2 draw their unlabeled budget.

The data is available at: https://github.com/geekeemoli/Semi-supervised-learning-unlabeled-data-selection.

[1] The original study of Olier et al. [14] reports 2,764 datasets. We obtained 2,970 from OpenML, the difference stemming from curation thresholds and the particular ChEMBL release.

[2] pXC50 is the negative base-10 logarithm of the XC50, which is a generic notation for the concentration of a compound that produces 50% of a maximal effect. XC50 may refer to concentration causing 50% inhibition (IC50), concentration causing 50% activation/effect value (EC50), or similar. For more details, see Olier et al. [14].

### 3.2 Evaluation strategy

For each of the 10 datasets we evaluate every semi-supervised method (self-training, co-training, and label propagation; Section 2) in combination with every unlabeled-data selection strategy (Random, Nearest, Furthest, and Mixed; Section 2.2), and unlabeled-set sizes $m \in \{100, 250, 500, 1000, 2500, 5000\}$, and compare it against a supervised random forest baseline.

*Cross-validation.* Performance is estimated by $k$-fold cross-validation over the labeled molecules of each dataset. Because the datasets vary widely in size, the number of folds is adapted to the amount of available labeled data: datasets with at least 100 labeled molecules use 10-fold cross-validation and datasets with between 30 and 99 molecules use 5-fold cross-validation. In every case, the folds are created with a fixed random seed, so the partition is identical across methods, strategies, and unlabeled-set sizes. This yields a paired comparison in which the data split is held constant, isolating the effect of the method from variation due to how the labeled data happens to be partitioned.

*Training, test, and unlabeled sets.* In each cross-validation iteration, the labeled molecules are divided into a training fold $\mathcal{L}_{\text{train}}$ and a test fold $\mathcal{L}_{\text{test}}$. The supervised baseline is trained on $\mathcal{L}_{\text{train}}$ alone. For the semi-supervised methods, an unlabeled set of size $m$ is selected from the external pool of 415,654 molecules (Section 3.1) using the chosen selection strategy and $\mathcal{L}_{\text{train}}$, and the method is trained on $\mathcal{L}_{\text{train}}$ together with this unlabeled set. Both models are then evaluated on the same test fold $\mathcal{L}_{\text{test}}$. To prevent information leakage, the unlabeled selection uses only the training fold as its reference for distance computations; the test fold is never consulted during selection, and the external pool excludes all molecules belonging to the dataset under consideration. The selection is repeated independently in every fold, since the training fold (and hence the reference set) changes from fold to fold.

*Metrics.* Predictive performance is measured by the root mean squared error (RMSE) between the predicted and the true pXC50 values of test fold molecules. For a given combination of method, selection strategy, and unlabeled-set size $m$, the RMSE is averaged across the cross-validation folds of a dataset to obtain a per-dataset estimate. To compare SSL performance across datasets, we express results as the percentage improvement over the supervised baseline. Improvement is defined as the relative reduction in RMSE achieved by an SSL model:

$$\mathit{\Delta}\text{RMSE}(\%) = 100 \cdot \frac{\text{RMSE}_{\text{sup}} - \text{RMSE}_{\text{SSL}}}{\text{RMSE}_{\text{sup}}}, \tag{6}$$

Positive values indicate improved performance of SSL model, while negative values indicate worse performance than the supervised random forest baseline.

Table 2: Best SSL configuration for each dataset. The table reports the SSL method, unlabeled-data selection strategy, and unlabeled-set size ($m$) yielding the lowest RMSE. $\Delta$RMSE denotes the improvement relative to the supervised random forest baseline.

| Dataset | #Molecules | $m$ | Method | Strategy | $\text{RMSE}_{\text{sup}}$ | $\text{RMSE}_{\text{ssl}}$ | $\Delta$RMSE(%) |
|---|---|---|---|---|---|---|---|
| CHEMBL1907606 | 30 | 5000 | Label prop | Mixed | 1.256 | 1.036 | 17.5 |
| CHEMBL4010 | 33 | 100 | Co-train | Nearest | 0.432 | 0.408 | 5.5 |
| CHEMBL4835 | 35 | 250 | Label prop | Nearest | 0.838 | 0.785 | 6.4 |
| CHEMBL4208 | 37 | 1000 | Label prop | Mixed | 0.828 | 0.722 | 12.8 |
| CHEMBL2457 | 70 | 2500 | Label prop | Mixed | 0.278 | 0.200 | 28.0 |
| CHEMBL5281 | 75 | 500 | Label prop | Random | 0.557 | 0.518 | 7.0 |
| CHEMBL5685 | 75 | 250 | Co-train | Mixed | 0.549 | 0.536 | 2.4 |
| CHEMBL3535 | 112 | 2500 | Self-train | Nearest | 0.613 | 0.604 | 1.4 |
| CHEMBL4698 | 264 | 500 | Self-train | Random | 0.695 | 0.685 | 1.4 |
| CHEMBL2434 | 525 | 100 | Self-train | Furthest | 0.639 | 0.635 | 0.5 |

*Hyperparameters.* All methods share a random forest base learner with 100 trees. The method-specific hyperparameters were fixed to sensible default values rather than tuned per dataset, in order to keep the comparison focused on the data-selection question and to avoid optimistic bias from per-dataset tuning. For self-training, an unlabeled prediction is accepted as a pseudo-label when the standard deviation of the tree predictions is at most 0.5 in pXC50 units, and at most 5 self-training rounds are performed. For co-training regression, 10 rounds are run, with the 10 most-agreed-upon molecules exchanged between the two views per round. For label propagation, a $k$-nearest-neighbor graph with $k = 10$ is used, the propagation runs for at most 50 iterations or until convergence, and an unlabeled molecule is retained as a pseudo-label only if its propagated value deviates from the prior mean by at least 0.1 in pXC50 units.

## 4 Results and Discussion

Due to the large number of experiments, per-dataset results are provided in Supplementary Figure S1 and Table S1, while in the main manuscript, we focus on aggregate analyses to facilitate interpretation and highlight the main trends.

### 4.1 Can Large Chemical Repositories Improve QSAR Models via SSL?

We begin by addressing the first research question: whether large chemical repositories can enhance QSAR models via SSL. We observe that, on each of the 10 datasets, at least one SSL configuration outperforms the supervised random forest baseline, with improvements ranging from 0.5% to 28% (Table 2). This demonstrates that SSL methods can successfully exploit unlabeled molecules extracted from large chemical repositories to improve QSAR models.

From Table 2, we can observe that label propagation achieved the largest gains in predictive performance and achieved the best performance in 5 out of 10 datasets. However, no SSL method consistently dominates across datasets, as label propagation, self-training, and co-training each achieve the best performance on at least two datasets, indicating complementarity among the methods. Second, the largest gains are observed on the smallest datasets (with 75 or fewer labeled molecules), suggesting that SSL is particularly beneficial when labeled data are scarce. Finally, the best-performing configurations are distributed across all unlabeled-data selection strategies and unlabeled-set sizes, motivating a more detailed investigation of these factors (Section 4.2).

Examining the complete set of SSL configurations (Table S1, Figure S1), we observe substantial heterogeneity in predictive performance across methods, unlabeled-data selection strategies, and unlabeled set size. Some datasets appear more conducive to specific SSL methods, such as `CHEMBL1907606` and `CHEMBL2457`, where a large proportion of the self-training and co-training configurations outperform the supervised baseline. Similarly, on `CHEMBL4208`, label propagation consistently achieves improvements. On the other hand, on datasets such as `CHEMBL2434`, `3535`, and `5685`, only a few specific configurations yield improvements, while many others degrade predictive performance. Although label propagation achieved the largest overall improvement on 5 out of 10 datasets, it is also the least robust method, exhibiting substantial performance degradations on several datasets (e.g., `CHEMBL3535`, `5281`, and `5685`).

The observed heterogeneity is consistent with established findings in the SSL literature [5]. SSL methods rely on assumptions about the relationship between the data distribution and labels, such as the smoothness assumption (i.e., nearby instances tend to share labels) and the manifold assumption (i.e., the data lie on a low-dimensional manifold). When these assumptions are violated, incorporating unlabeled data may degrade predictive performance.

While the best-performing configurations (Table 2) demonstrate the potential of large chemical repositories to improve QSAR models, these results do not provide the means to select an appropriate SSL configuration in practice. Determining which SSL method is best suited to a given dataset remains an open research question in the SSL field and is the subject of ongoing work on algorithm selection and meta-learning for SSL [17]. In parallel, another line of research focuses on safe SSL methods that are guaranteed to perform no worse than their supervised counterparts [10]. Although these questions are highly relevant, developing methods that can reliably identify beneficial SSL configurations and consistently avoid performance degradation is beyond the scope of this study. We nevertheless return to this issue in the next section, where we discuss some practical guidelines for unlabeled-data selection.

### 4.2 How Does Unlabeled Data Selection Affect SSL Performance?

We next examine the second research question: how does unlabeled data selection affect the performance of SSL models? To this end, we aggregate the predictive performance results across datasets and report the average relative improvement over the supervised baseline (Figure 2a).

The three methods exhibit different preferences with respect to unlabeled data selection strategies. Self-training generally benefits from Random and Furthest strategies and achieves improvements across most unlabeled set sizes ($m$). Co-training yields improvements only for small unlabeled sets, specifically with the Nearest strategy at $m = 100$ and the Furthest strategy at $m = 250$. While the performance of the Nearest strategy deteriorates rapidly as $m$ increases, the Furthest strategy remains more stable across different unlabeled-set sizes. Finally, label propagation also achieves improvements for small unlabeled sets with Random ($m = 100$) and Nearest ($m = 250$) strategies.

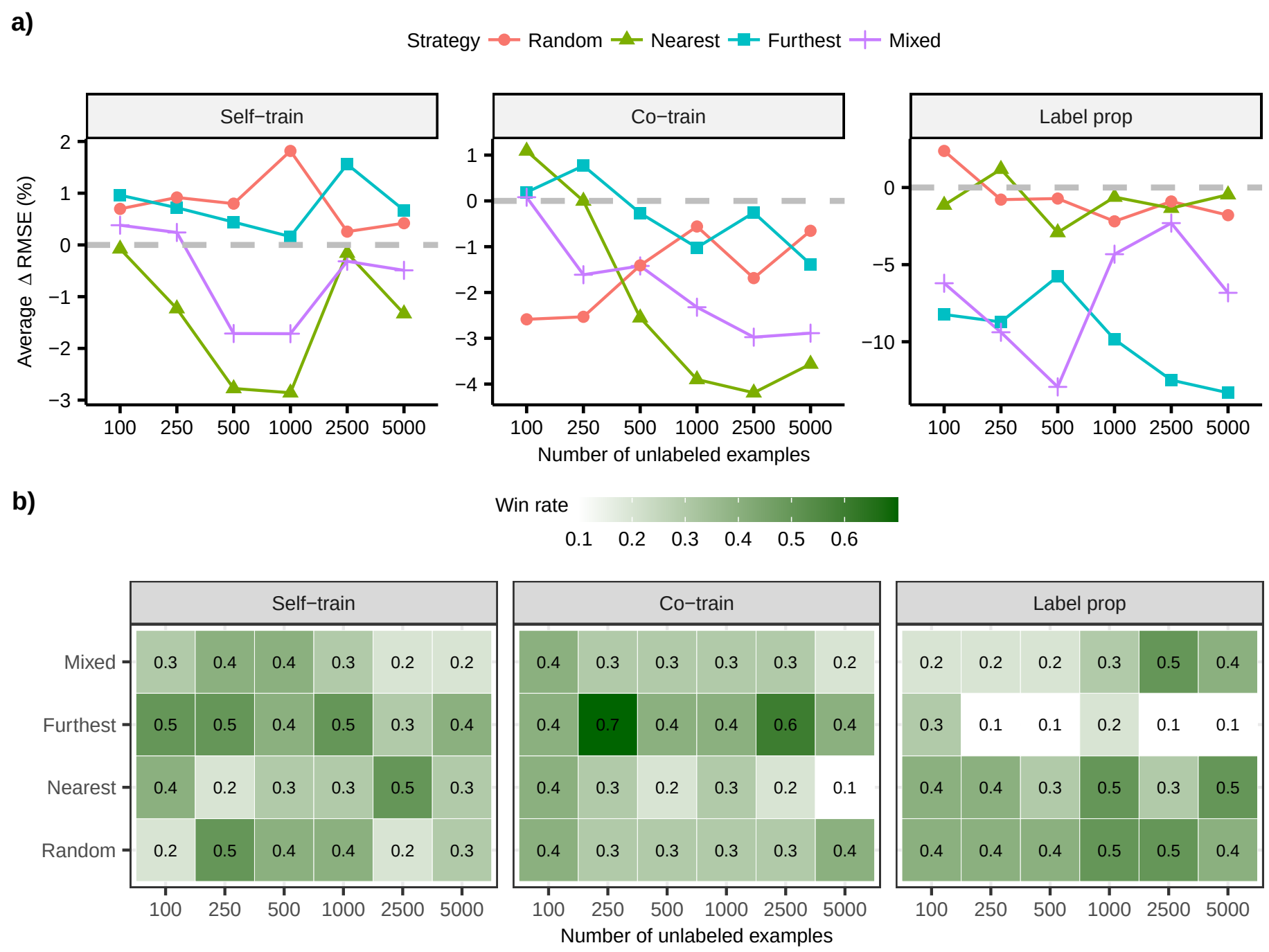

Fig. 2: Effect of SSL method, unlabeled-data selection strategy, and unlabeled-set size. (a) Average relative improvement (average $\Delta$RMSE) over the supervised baseline across the 10 QSAR datasets. (b) Win rate, i.e., the proportion of datasets on which a given SSL configuration improves upon the supervised baseline.

In contrast to self-training and co-training, label propagation favours the Nearest strategy, whereas the Furthest strategy consistently leads to substantial performance degradation. A possible explanation is that label propagation exploits the manifold assumption by propagating information through a similarity graph. Consequently, unlabeled molecules that are close to the labeled data may better preserve the local graph structure, whereas chemically distant molecules can distort the graph and hinder label propagation. On the other hand, self-training and co-training rely on iteratively generated pseudo-labels and may benefit from expanded coverage of the chemical space beyond the labeled training set.

Table 3: Average ranks for SSL methods, unlabeled-data selection strategies, and unlabeled-set sizes. Lower ranks indicate better performance. The best performer is marked in bold.

| SSL method | | Selection strategy | | Unlabeled-set size | |
|---|---|---|---|---|---|
| Method | Rank | Strategy | Rank | $m$ | Rank |
| Self-train | **1.81** | Random | 2.31 | 100 | **3.08** |
| Co-train | 2.09 | Nearest | 2.74 | 250 | 3.27 |
| Label prop | 2.10 | Furthest | **2.29** | 500 | 3.50 |
| | | Mixed | 2.67 | 1000 | 3.64 |
| | | | | 2500 | 3.61 |
| | | | | 5000 | 3.90 |

Overall, a common pattern emerges across all methods: increasing the number of unlabeled examples does not consistently improve performance. Instead, the best results are often obtained using relatively small unlabeled sets, suggesting that the composition of the unlabeled data may be more important than its size.

Figure 2a shows the average magnitude of performance improvements of different SSL configurations. To derive practical recommendations, we additionally analyse the win rate across datasets, which indicates how frequently a given configuration outperforms the supervised baseline (Figure 2b). Self-training combined with the Furthest strategy appears to be the most robust configuration, outperforming the baseline on up to 50% of datasets while also yielding consistently positive average improvements across different unlabeled-set sizes. Co-training achieves the highest observed win rate (70%) when combined with the Furthest strategy and $m = 250$. However, substantial performance degradations on some datasets (e.g., `CHEMBL2434` and `4698`; Figure S1) result in less consistent average improvements. Finally, label propagation achieves win rates of 40-50% when combined with the Nearest and Random strategies, although its average performance remains highly variable across datasets.

To provide an overall summary of the experimental results, Table 3 reports average ranks of SSL methods, unlabeled-data selection strategies, and unlabeled-set sizes. Self-training achieves the best average rank among the SSL methods, although the differences in rank between methods are relatively small. Among the selection strategies, Furthest obtains the best average rank, followed by Random, while Nearest performs the worst. Smaller unlabeled sets consistently rank higher than larger ones, with $m = 100$ yielding the best overall performance and $m = 5000$ the worst. Notably, rank differences are more pronounced across selection strategies and unlabeled-set sizes than across SSL methods, suggesting that the choice of unlabeled data may have a greater impact on predictive performance than the choice of SSL algorithm. These findings reinforce our conclusion that careful selection of unlabeled data is more important than simply increasing its quantity.

## 5 Conclusions

In this study, we investigated whether large chemical repositories can be exploited to improve Quantitative Structure–Activity Relationship (QSAR) models through semi-supervised learning (SSL) and how the selection of unlabeled data affects SSL performance. Across 10 QSAR datasets, we found that unlabeled molecules extracted from the ChEMBL database can indeed improve predictive performance, with at least one SSL configuration outperforming the supervised baseline on every dataset. While the benefits of SSL strongly depend on the choice of method, the composition of the unlabeled set, and the number of unlabeled examples, the most robust improvements over supervised random forests were generally achieved by self-training combined with Furthest unlabeled-data selection strategy and a small unlabeled set of up to 250 molecules.

Our results demonstrate that unlabeled-data selection is a critical factor in QSAR modelling with SSL. In particular, selecting a relatively small number of chemically distant molecules often leads to more reliable improvements than using large collections of unlabeled compounds. This finding challenges the common assumption that SSL primarily benefits from increasing the amount of unlabeled data and suggests that the composition of the unlabeled set may be more important than its size. From a practical perspective, this implies that practitioners may not need to curate or process very large unlabeled collections to benefit from SSL. Instead, carefully selecting a relatively small set of informative molecules may achieve better results while reducing computational costs.

These findings should be considered in light of the limitations of this study. Even though our analysis yielded practical recommendations for applying SSL to QSAR modelling, further work on meta-learning and safe semi-supervised learning is needed to improve the reliability and usability of SSL in practice. Additional limitations include a limited number of datasets, a single application domain, fixed SSL hyperparameters, and the use of relatively simple unlabeled-data selection strategies. Future work will expand the collection of QSAR datasets and consider other application domains, investigate additional SSL methods and hyperparameter settings, and develop more sophisticated approaches for selecting informative unlabeled molecules from large repositories.

**Acknowledgments.** We acknowledge the support of the Slovenian Research and Innovation Agency via the research project "AI for Science" (grant number GC-0001).

**Disclosure of Interests.** The authors have no competing interests to declare that are relevant to the content of this article.

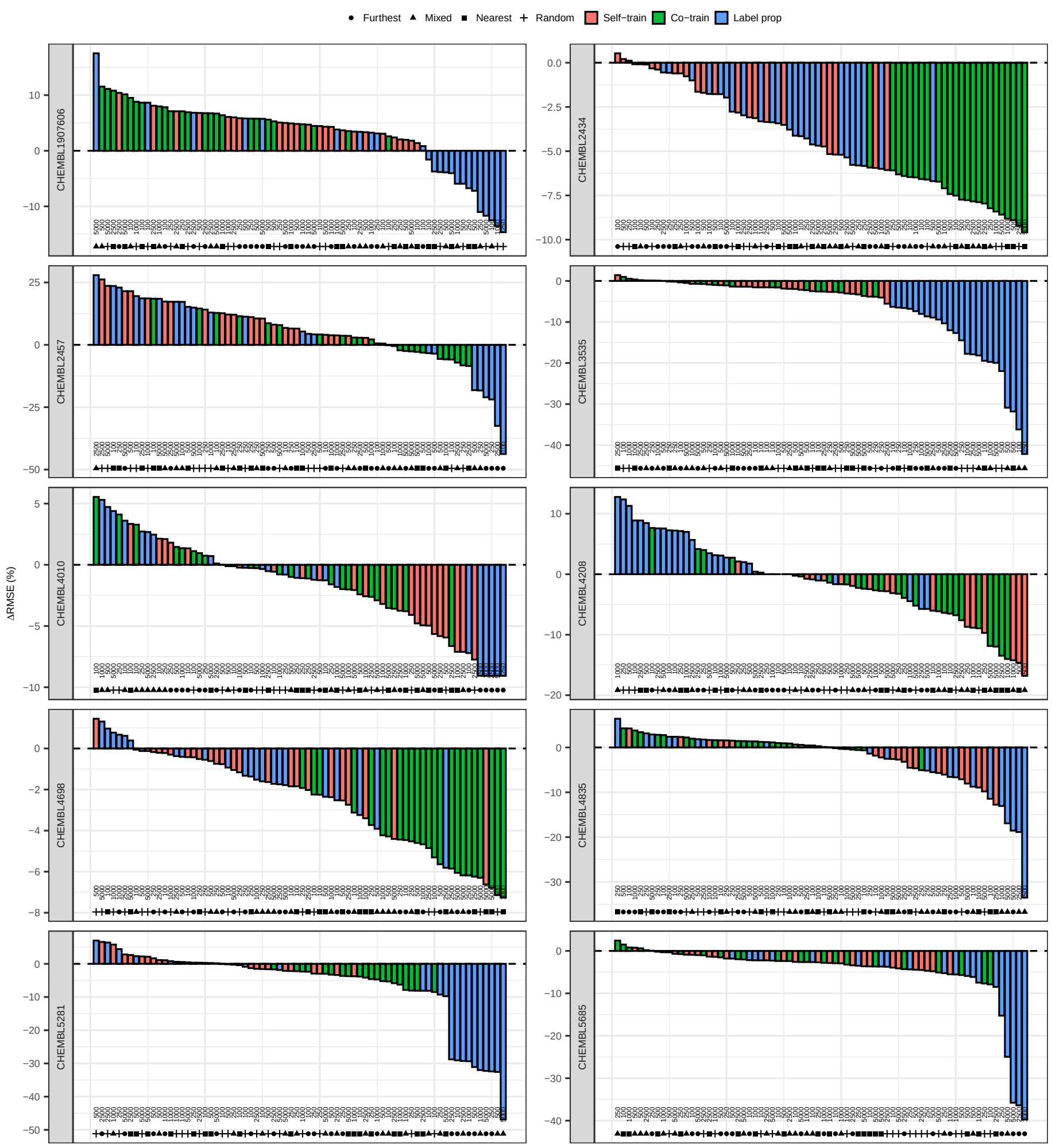

Supplementary Figure S1: Ranked improvements ($\Delta$RMSE) of all SSL configurations relative to the supervised baseline for each QSAR dataset. Configurations are ordered from best to worst within each dataset. Colors represent SSL methods and symbols represent unlabeled-data selection strategies; numbers indicate the size of the selected unlabeled set ($m$).

Supplementary Table S1: Results (RMSE) for different SSL algorithms and unlabeled data selection strategies across 10 ChEMBL datasets and varying numbers of unlabeled molecules $m \in \{100, 250, 500, 1000, 2500, 5000\}$. **Bold** denotes the best performance per $m$ and underline that the SSL algorithm outperformed the supervised random forest baseline.

| Method | Strategy | 100 | 250 | 500 | 1000 | 2500 | 5000 | 100 | 250 | 500 | 1000 | 2500 | 5000 |
|---|---|---|---|---|---|---|---|---|---|---|---|---|---|
| | | CHEMBL1907606 | | | | | | CHEMBL4010 | | | | | |
| Baseline | - | 1.256 | 1.256 | 1.256 | 1.256 | 1.256 | 1.256 | 0.432 | 0.432 | 0.432 | 0.432 | 0.432 | 0.432 |
| Self-train | Random | 1.2 | 1.192 | 1.171 | 1.18 | 1.181 | 1.202 | 0.434 | 0.449 | 0.457 | 0.445 | 0.465 | 0.44 |
| Self-train | Nearest | 1.239 | 1.226 | 1.231 | 1.194 | 1.154 | 1.21 | 0.417 | 0.436 | 0.453 | 0.462 | 0.457 | 0.452 |
| Self-train | Furthest | 1.213 | 1.182 | 1.195 | 1.202 | 1.125 | 1.193 | 0.426 | 0.424 | 0.437 | 0.433 | 0.448 | 0.456 |
| Self-train | Mixed | 1.218 | 1.23 | 1.214 | 1.197 | 1.167 | 1.233 | 0.422 | 0.422 | 0.432 | 0.443 | 0.462 | 0.453 |
| Co-train | Random | 1.223 | **1.167** | 1.189 | **1.145** | 1.169 | 1.116 | 0.427 | **0.414** | 0.433 | 0.435 | 0.443 | 0.44 |
| Co-train | Nearest | 1.147 | **1.167** | 1.186 | 1.176 | **1.12** | 1.129 | **0.408** | 0.436 | 0.442 | 0.448 | 0.46 | 0.447 |
| Co-train | Furthest | 1.158 | 1.183 | 1.183 | 1.196 | 1.171 | 1.2 | 0.433 | 0.428 | 0.425 | 0.426 | 0.431 | 0.427 |
| Co-train | Mixed | **1.137** | 1.212 | **1.111** | 1.156 | 1.171 | 1.172 | 0.417 | 0.438 | 0.436 | 0.44 | 0.447 | 0.444 |
| Label prop | Random | 1.147 | 1.44 | 1.331 | 1.426 | 1.304 | 1.403 | 0.432 | 0.435 | 0.432 | 0.433 | 0.437 | **0.413** |
| Label prop | Nearest | 1.276 | 1.341 | 1.346 | 1.208 | 1.303 | 1.307 | 0.437 | 0.436 | 0.433 | 0.439 | 0.434 | 0.428 |
| Label prop | Furthest | 1.245 | 1.217 | 1.183 | 1.215 | 1.17 | 1.184 | 0.463 | 0.471 | 0.471 | 0.471 | 0.471 | 0.471 |
| Label prop | Mixed | 1.413 | 1.394 | 1.305 | 1.33 | 1.213 | **1.036** | 0.416 | 0.42 | **0.411** | **0.409** | **0.421** | 0.42 |
| | | CHEMBL4835 | | | | | | CHEMBL4208 | | | | | |
| Baseline | - | 0.838 | 0.838 | 0.838 | 0.838 | 0.838 | 0.838 | 0.828 | 0.828 | 0.828 | 0.828 | 0.828 | 0.828 |
| Self-train | Random | 0.854 | 0.826 | 0.898 | 0.841 | 0.877 | 0.921 | 0.828 | 0.811 | 0.837 | 0.842 | 0.849 | 0.854 |
| Self-train | Nearest | 0.825 | 0.889 | 0.876 | 0.913 | 0.945 | 0.86 | 0.878 | 0.909 | 0.949 | 0.901 | 0.826 | 0.851 |
| Self-train | Furthest | **0.803** | 0.819 | 0.834 | 0.825 | 0.865 | 0.906 | 0.828 | 0.828 | 0.828 | 0.828 | 0.835 | 0.825 |
| Self-train | Mixed | 0.841 | 0.825 | 0.886 | 0.857 | 0.84 | 0.861 | 0.83 | 0.9 | 0.967 | 0.946 | 0.834 | 0.844 |
| Co-train | Random | 0.831 | 0.83 | 0.827 | 0.825 | 0.843 | 0.82 | 0.831 | 0.842 | 0.84 | 0.902 | 0.884 | 0.871 |
| Co-train | Nearest | 0.844 | 0.81 | 0.828 | 0.829 | 0.835 | 0.839 | 0.882 | 0.861 | 0.851 | 0.944 | 0.94 | 0.927 |
| Co-train | Furthest | 0.815 | 0.815 | **0.803** | **0.807** | 0.826 | 0.827 | 0.765 | 0.795 | 0.806 | 0.85 | 0.794 | 0.847 |
| Co-train | Mixed | 0.835 | 0.881 | 0.833 | 0.831 | 0.822 | 0.827 | 0.881 | 0.855 | 0.848 | 0.879 | 0.891 | 0.926 |
| Label prop | Random | 0.837 | 0.812 | 0.823 | 0.828 | 0.824 | 0.842 | **0.755** | **0.726** | **0.735** | 0.803 | 0.765 | 0.799 |
| Label prop | Nearest | 0.934 | **0.785** | 0.86 | 0.836 | **0.818** | **0.814** | 0.802 | 0.769 | 0.758 | 0.77 | **0.755** | 0.805 |
| Label prop | Furthest | 0.818 | 0.85 | 0.885 | 0.894 | 0.894 | 0.994 | 0.768 | 0.837 | 0.842 | 0.865 | 0.876 | 0.876 |
| Label prop | Mixed | 0.912 | 0.882 | 1.119 | 0.948 | 0.997 | 0.98 | 0.768 | 0.812 | 0.814 | **0.722** | 0.781 | **0.766** |
| | | CHEMBL2457 | | | | | | CHEMBL5281 | | | | | |
| Baseline | - | 0.278 | 0.278 | 0.278 | 0.278 | 0.278 | 0.278 | 0.557 | 0.557 | 0.557 | 0.557 | 0.557 | 0.557 |
| Self-train | Random | 0.244 | 0.238 | **0.218** | **0.226** | 0.205 | **0.212** | 0.557 | 0.573 | 0.566 | 0.553 | 0.565 | 0.544 |
| Self-train | Nearest | 0.259 | 0.247 | 0.27 | 0.267 | 0.242 | 0.248 | 0.559 | 0.569 | 0.555 | 0.577 | 0.542 | 0.545 |
| Self-train | Furthest | 0.266 | 0.259 | 0.255 | 0.267 | 0.23 | 0.218 | 0.564 | 0.573 | 0.547 | 0.551 | **0.52** | **0.541** |
| Self-train | Mixed | 0.259 | 0.248 | 0.268 | 0.276 | 0.244 | 0.229 | 0.565 | 0.558 | 0.55 | **0.524** | 0.575 | 0.568 |
| Co-train | Random | 0.294 | 0.3 | 0.266 | 0.237 | 0.267 | 0.256 | 0.604 | 0.601 | 0.557 | 0.57 | 0.555 | 0.554 |
| Co-train | Nearest | 0.226 | 0.246 | 0.286 | 0.294 | 0.301 | 0.285 | 0.578 | 0.582 | 0.602 | 0.591 | 0.589 | 0.577 |
| Co-train | Furthest | 0.276 | 0.253 | 0.278 | 0.271 | 0.27 | 0.284 | 0.57 | 0.577 | 0.575 | 0.583 | 0.566 | 0.556 |
| Co-train | Mixed | 0.242 | 0.269 | 0.293 | 0.284 | 0.297 | 0.285 | 0.586 | 0.569 | 0.586 | 0.6 | 0.58 | 0.573 |
| Label prop | Random | 0.223 | 0.246 | 0.265 | 0.242 | 0.266 | 0.236 | 0.561 | **0.532** | **0.518** | 0.554 | 0.521 | 0.567 |
| Label prop | Nearest | **0.212** | **0.214** | 0.235 | 0.263 | 0.226 | 0.226 | **0.555** | 0.555 | 0.544 | 0.552 | 0.602 | 0.611 |
| Label prop | Furthest | 0.399 | 0.338 | 0.287 | 0.287 | 0.368 | 0.336 | 0.718 | 0.737 | 0.73 | 0.72 | 0.717 | 0.736 |
| Label prop | Mixed | 0.279 | 0.328 | 0.328 | 0.23 | **0.2** | 0.23 | 0.602 | 0.608 | 0.738 | 0.735 | 0.719 | 0.817 |
| | | CHEMBL5685 | | | | | | CHEMBL3535 | | | | | |
| Baseline | - | 0.549 | 0.549 | 0.549 | 0.549 | 0.549 | **0.549** | 0.613 | **0.613** | **0.613** | 0.613 | 0.613 | **0.613** |
| Self-train | Random | 0.571 | 0.575 | 0.573 | 0.557 | 0.574 | 0.572 | 0.628 | 0.618 | 0.624 | **0.61** | 0.647 | 0.633 |
| Self-train | Nearest | 0.569 | 0.564 | 0.562 | 0.565 | 0.556 | 0.569 | 0.623 | 0.632 | 0.625 | 0.611 | **0.604** | 0.632 |
| Self-train | Furthest | 0.554 | 0.559 | 0.575 | 0.579 | 0.568 | 0.554 | 0.623 | 0.638 | 0.636 | 0.622 | 0.622 | 0.622 |
| Self-train | Mixed | 0.563 | 0.554 | **0.545** | 0.565 | 0.561 | 0.553 | 0.623 | 0.629 | 0.619 | 0.617 | 0.625 | 0.626 |
| Co-train | Random | 0.592 | 0.591 | 0.58 | 0.567 | 0.555 | 0.577 | **0.607** | 0.637 | 0.63 | 0.623 | 0.629 | 0.62 |
| Co-train | Nearest | **0.541** | 0.553 | 0.55 | **0.545** | 0.56 | 0.569 | 0.629 | 0.614 | 0.627 | 0.621 | 0.629 | 0.635 |
| Co-train | Furthest | 0.565 | 0.549 | 0.557 | 0.564 | 0.573 | 0.563 | 0.615 | **0.613** | **0.613** | 0.619 | 0.612 | 0.617 |
| Co-train | Mixed | 0.562 | **0.536** | 0.546 | 0.563 | 0.562 | 0.56 | 0.623 | 0.614 | 0.618 | 0.613 | 0.612 | 0.616 |
| Label prop | Random | 0.561 | 0.573 | 0.551 | 0.58 | 0.569 | 0.562 | 0.653 | 0.702 | 0.748 | 0.722 | 0.723 | 0.735 |
| Label prop | Nearest | 0.561 | 0.568 | 0.581 | 0.59 | 0.596 | 0.569 | 0.732 | 0.653 | 0.808 | 0.655 | 0.687 | 0.666 |
| Label prop | Furthest | 0.561 | 0.633 | 0.559 | 0.767 | 0.749 | 0.745 | 0.652 | 0.658 | 0.671 | 0.662 | 0.676 | 0.691 |
| Label prop | Mixed | 0.563 | 0.578 | 0.583 | 0.551 | **0.548** | 0.686 | 0.835 | 0.871 | 0.802 | 0.734 | 0.668 | 0.724 |
| | | CHEMBL4698 | | | | | | CHEMBL2434 | | | | | |
| Baseline | - | 0.695 | **0.695** | 0.695 | 0.695 | 0.695 | 0.695 | 0.639 | 0.639 | 0.639 | **0.639** | **0.639** | **0.639** |
| Self-train | Random | 0.698 | 0.703 | **0.685** | 0.697 | 0.697 | 0.696 | 0.641 | **0.638** | **0.637** | 0.644 | 0.659 | 0.661 |
| Self-train | Nearest | 0.708 | 0.699 | 0.713 | 0.719 | 0.709 | 0.741 | 0.639 | 0.643 | 0.65 | 0.657 | 0.669 | 0.678 |
| Self-train | Furthest | 0.697 | 0.696 | 0.7 | 0.698 | 0.714 | 0.702 | **0.635** | 0.641 | 0.639 | 0.649 | 0.66 | 0.677 |
| Self-train | Mixed | 0.708 | 0.7 | 0.699 | 0.712 | 0.706 | 0.726 | 0.639 | 0.643 | 0.65 | 0.659 | 0.672 | 0.672 |
| Co-train | Random | 0.711 | 0.708 | 0.745 | 0.732 | 0.729 | 0.739 | 0.681 | 0.678 | 0.686 | 0.693 | 0.698 | 0.694 |
| Co-train | Nearest | 0.711 | 0.721 | 0.727 | 0.746 | 0.738 | 0.742 | 0.687 | 0.688 | 0.696 | 0.695 | 0.69 | 0.7 |
| Co-train | Furthest | 0.726 | 0.726 | 0.734 | 0.728 | 0.738 | 0.737 | 0.681 | 0.68 | 0.679 | 0.68 | 0.677 | 0.682 |
| Co-train | Mixed | 0.717 | 0.727 | 0.725 | 0.724 | 0.736 | 0.738 | 0.68 | 0.691 | 0.688 | 0.684 | 0.689 | 0.689 |
| Label prop | Random | 0.702 | 0.699 | 0.696 | **0.69** | **0.691** | **0.686** | 0.656 | 0.666 | 0.66 | 0.66 | 0.658 | 0.676 |
| Label prop | Nearest | **0.688** | 0.705 | 0.711 | 0.718 | 0.735 | 0.692 | 0.661 | 0.676 | 0.672 | 0.665 | 0.668 | 0.663 |
| Label prop | Furthest | 0.713 | 0.704 | 0.698 | **0.69** | 0.707 | 0.707 | 0.65 | 0.642 | 0.645 | 0.65 | 0.642 | 0.651 |
| Label prop | Mixed | 0.695 | 0.706 | 0.722 | 0.706 | 0.698 | 0.707 | 0.665 | 0.676 | 0.682 | 0.677 | 0.673 | 0.669 |